\documentclass[conference]{IEEEtran}

\usepackage{subfigure}
\usepackage{hyperref}       
\usepackage{algorithm}
\usepackage[flushleft]{threeparttable}
\usepackage{multirow}
\usepackage{algorithmic}
\usepackage{booktabs}       
\usepackage{nicefrac}       
\usepackage{microtype}      
\usepackage{amsthm, amsmath, amsfonts, amssymb}
\usepackage{mathrsfs,mathtools}
\usepackage{bbm}
\usepackage{bm}
\usepackage{xcolor}
\usepackage{stackengine}
\usepackage{nicefrac}       
\usepackage{microtype}      

\newcommand*{\eg}{\emph{e.g.}{}}
\newcommand*{\ie}{\emph{i.e.}{}}

\newtheorem{prop}{Proposition}

\newtheorem{rmk}{Remark}
\IEEEoverridecommandlockouts
\usepackage{cite}
\usepackage{amsmath,amssymb,amsfonts}
\usepackage{algorithmic}
\usepackage{graphicx}
\usepackage{textcomp}
\usepackage{xcolor}
\def\BibTeX{{\rm B\kern-.05em{\sc i\kern-.025em b}\kern-.08em
    T\kern-.1667em\lower.7ex\hbox{E}\kern-.125emX}}
\begin{document}

\title{Uncertainty-Aware Robust Learning \\on Noisy Graphs
}

\author{
\IEEEauthorblockN{Shuyi Chen}
\IEEEauthorblockA{\textit{Carnegie Mellon University} \\
Pittsburgh, PA, USA \\
shuyic@andrew.cmu.edu}
\and
\IEEEauthorblockN{Kaize Ding}
\IEEEauthorblockA{\textit{Northwestern University} \\
Evanston, IL, USA \\
kaize.ding@northwestern.edu}
\and
\IEEEauthorblockN{Shixiang Zhu}
\IEEEauthorblockA{\textit{Carnegie Mellon University} \\
Pittsburgh, PA, USA \\
shixianz@andrew.cmu.edu}
}


\maketitle

\begin{abstract}
  Graph neural networks (GNNs) have excelled in various graph learning tasks, particularly node classification. However, their performance is often hampered by noisy measurements in real-world graphs, which can corrupt critical patterns in the data. To address this, we propose a novel uncertainty-aware graph learning framework inspired by distributionally robust optimization.
  Specifically, we use a graph neural network-based encoder to embed the node features and find the optimal node embeddings by minimizing the worst-case risk through a minimax formulation. Such an uncertainty-aware learning process leads to improved node representations and a more robust graph predictive model that effectively mitigates the impact of uncertainty arising from data noise. 
  Our experimental results demonstrate superior predictive performance over baselines across noisy scenarios. 
\end{abstract}


\section{Introduction}
The field of graph learning has witnessed significant advancements in recent years, fueled by the remarkable performance of graph neural networks (GNNs) \cite{gcn, gat}. 
GNNs leverage message-passing techniques to enable efficient information exchange among nodes in a graph, leading to improved embeddings. 
Among the various graph-based learning tasks, node classification, especially in a semi-supervised setting, stands out as a prominent and widely applicable problem that has greatly benefited from GNNs. The objective of semi-supervised node classification is to learn high-quality node embeddings and make predictions on unlabeled nodes in a graph with only a small subset of nodes labeled.

However, GNNs, like other deep neural networks, are highly sensitive to noise, including inaccuracies in both node features and graph structure \cite{li2022recent, zhu2021survey, 10.1145/3437963.3441734}. This issue is particularly problematic in low-data settings, where a limited number of labeled nodes are available, further degrading the model's performance \cite{ding2022toward}. For example, in social networks, new users may inconsistently express preferences or engage with content due to limited options, resulting in noisy data. Such noise introduces uncertainty and misleads the model, as the GNN relies on these noisy observations to learn and generalize preferences.

To tackle this, we introduce a novel uncertainty-aware framework, Distributionally Robust Graph Learning (DRGL), leveraging Distributionally Robust Optimization (DRO) \cite{duchi2021learning, shalev2016minimizing, NEURIPS2018_a08e32d2, zhu2022distributionally}. By modeling uncertainty with a Wasserstein ball \cite{NEURIPS2018_a08e32d2, zhang2021robust}, our approach seeks the least favorable distribution (LFD) for worst-case risk minimization. Furthermore, the resulting minimax solution also provides a means to estimate the potential uncertainty associated with predictions, which is particularly valuable in high-stakes scenarios where incorrect decisions can have severe consequences \cite{ABDAR2021243,begoli2019need,ryu2019bayesian,zhang2019bayesian}. 

\begin{figure}[!t]
  \centering
  \includegraphics[width=0.7\linewidth]{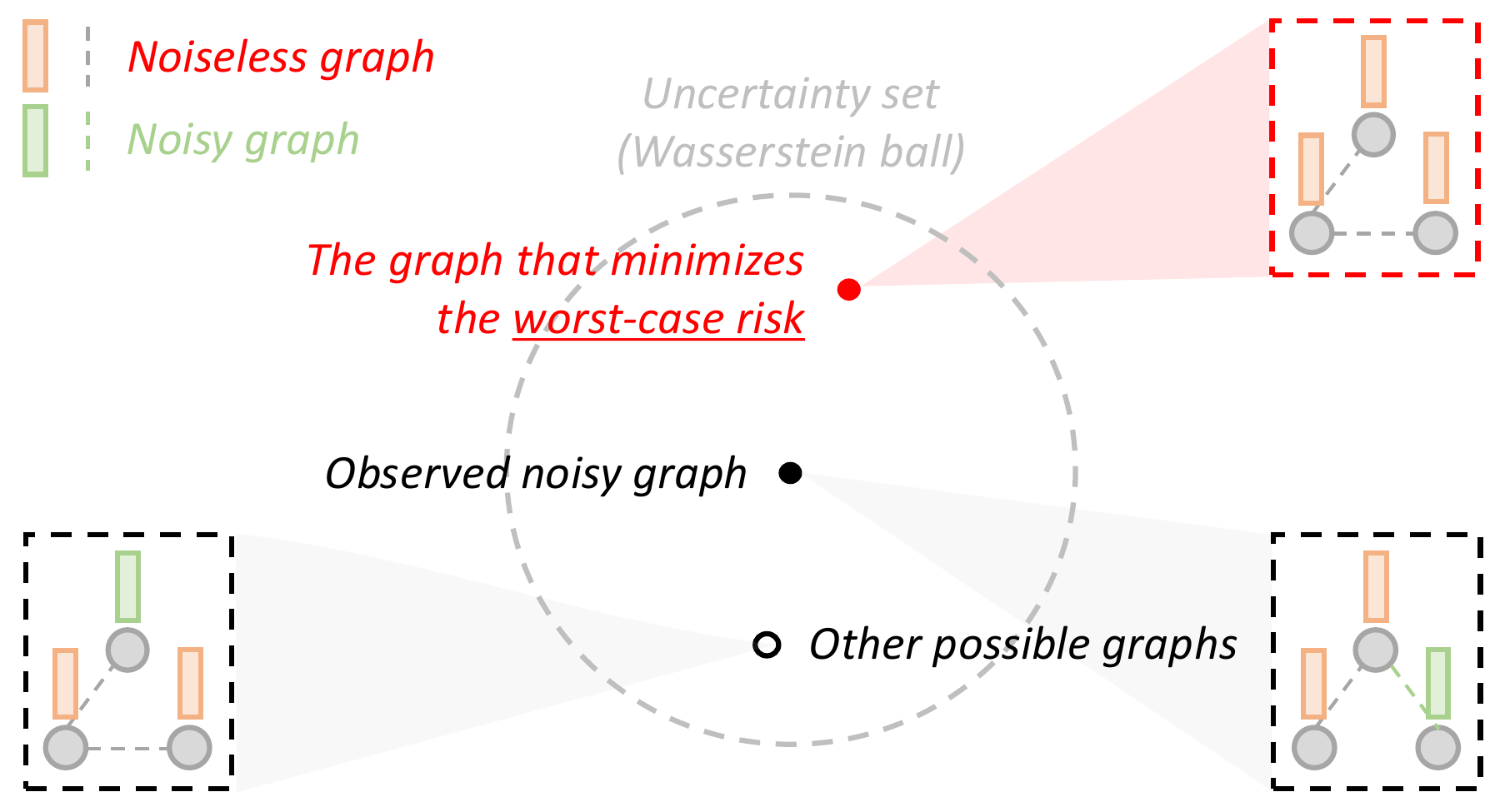}
  \caption{
  An illustration of the uncertainty set in our proposed framework. The goal is to search for the graph distribution that minimizes the worst-case risk. 
  }
  \label{fig:uncertainty_set_exp}
\vspace{-.2in}
\end{figure}

\section{Related Work}

Graph Neural Networks (GNNs) have proven highly effective for graph-structured data \cite{gcn,gat}, but they are sensitive to noisy features and graph structures \cite{li2022recent,10.1145/3437963.3441734}. To improve robustness, several methods have been proposed. Structure learning approaches aim to denoise graph structures, leveraging techniques such as graph structure learners \cite{NEURIPS2020_e05c7ba4, 10.1145/3394486.3403049}, while other works focus on noise-resistant aggregation for noisy or missing node features \cite{8970731,rossi2022unreasonable}. However, these methods typically require task-specific designs, limiting their general applicability.

Graph attack defenses also contribute to this domain. For example, GCN-SVD \cite{entezari2020all} reduces perturbation impact through low-rank approximation, while ProGNN \cite{jin2020graph} promotes robust learning by incorporating graph properties like low-rank and sparsity. RGCN \cite{zhu2019robust} employs Gaussian distributions to mitigate adversarial attacks. Although effective, these approaches are not extensively tested in low-data or random noise settings.

Our approach introduces a unified framework based on Distributionally Robust Optimization (DRO) \cite{duchi2021learning, shalev2016minimizing}. Unlike previous DRO methods designed for graph-based tasks \cite{zhang2021robust, sadeghi2021distributionally} that alternate between model parameters and distribution updates, our end-to-end approach uses gradient-based learning, providing robustness against both feature and structure noise. Furthermore, our framework enables uncertainty quantification under Least Favorable Distributions (LFDs), which is crucial in uncertainty-sensitive applications like molecular classification \cite{ABDAR2021243, ryu2019bayesian}.

\section{Uncertainty-aware Graph Learning}
\label{sec:method}

\subsection{Problem Setup}
\label{sec:setup}
Consider an attributed graph $(\mathcal{I}, \mathcal{E})$, where $\mathcal{I} = \{i=1, \dots, n\}$ represents the set of $n$ nodes, and $\mathcal{E} = \{(i,j), i, j \in \mathcal{I}\}$ represents the set of edges connecting the nodes.
Each node $i$ is associated with a $d$-dimensional feature vector $x_i \in \mathbb{R}^d$.
The collection of all node features is denoted as $X = [x_1, x_2, \dots, x_n] \in \mathbb{R}^{n \times d}$.
To describe the graph more generally, we can represent it as $\mathcal{G} = (X, A)$, where $A = (a_{ij}) \in \{0, 1\}^{n \times n}$ 
is the adjacency matrix, which provides information about the connectivity between nodes. If $(i, j) \in \mathcal{E}$, then $a_{ij} = 1$; otherwise, $a_{ij} = 0$.
Each node can be assigned one of the discrete labels $y \in \{m = 1, \dots, M\}$. 
In the context of semi-supervised node classification, we have access to labels for only a subset of nodes, which we denote as $y_{o} = [y_1, \dots, y_{n'}]^\top$, where $n'$ represents the number of observed nodes. The remaining nodes have no assigned labels and are denoted as $y_u = [y_{n'+1}, \dots, y_n]$.
In our setting, the assigned labels are assumed to be correct, but there might be errors or inaccuracies in the observed edges or node features due to noisy measurements.
Our objective is to accurately predict the labels $y_u$ for these unobserved nodes in the graph.

\subsection{Node Embedding}
We use a graph encoder to embed the node features, including both the nodal information and the corresponding graph structure. 
We emphasize that our framework is not tied to any specific graph model and offers flexibility in selecting graph encoders. In this study, we use Graph Convolutional Networks (GCNs) \cite{gcn,Li_Han_Wu_2018} and Graph Attention Networks (GATs) \cite{gat} to encode both the node feature and the graph topology. For each node $i$, the graph encoder functions as a nonlinear transformation, denoted as $\phi_\theta$, taking the nodal features and the corresponding graph topology as input and returning their node embeddings, denoted by $\xi \in \Xi$. Formally, 
\[
    \phi_\theta(\cdot, \mathcal{G}):~\mathcal{I} \rightarrow \Xi. 
\]
Here $\theta \in \Theta$ denotes the parameters specific to the model used in our framework.


It is important to note that the node embeddings obtained through the minimization of standard loss functions (\eg, cross-entropy loss) may not accurately capture the key feature patterns of the graph in our problem setting. 
The presence of noise in the observed graph $\mathcal{G}$, compounded by the scarcity of labeled nodes, can lead to overfitting on noisy data, misguided gradients, and poor generalization.

\subsection{Distributionally Robust Graph Learning}
\begin{figure}
  \centering
  \includegraphics[width=0.7\linewidth]{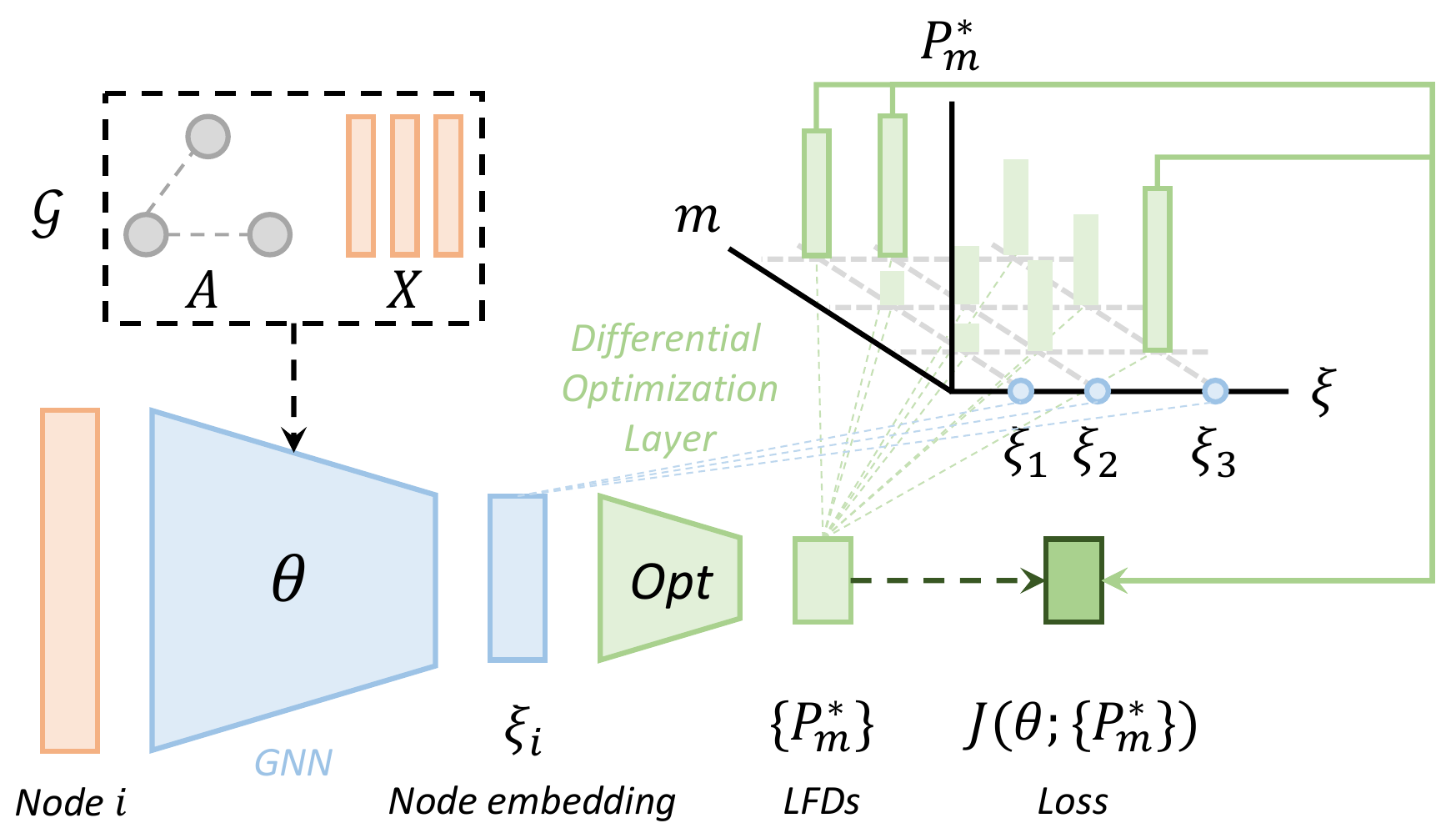}
  \caption{
  The architecture of the proposed framework consists of two cohesive modules: (1) a graph encoder parameterized by $\theta$, which produces the node embedding $\xi$ given the graph information $\mathcal{G}$; (2) a differential optimization layer, which generates the corresponding least favorable distributions (LFDs) $\{P_m^*\}$ for $\xi$ by solving the convex optimization defined in \eqref{eq:reformulation}. The loss measures the total margin, obtained by summing up $\max_{1\le m \le M} P_m^*(\xi)$ across node embeddings. 
  }
  \label{fig:architecture}
\end{figure}

To address these challenges, 
we propose a graph learning framework that improves the node embeddings, resulting in more robust predictive performance, particularly when confronted with noisy data. Figure~\ref{fig:architecture} summarizes the architecture of the proposed model.
Our approach assumes that node embeddings share the same label $m$ adhere to an underlying distribution ($\xi_i \sim P_m \in \mathcal{P}_m, \forall i: y_i = m$) within an uncertainty set $\mathcal{P}_m$ encompassing all potential distributions $P_m$.
However, obtaining a direct observation of this distribution is challenging due to inaccuracies in the nodal or topological information, and any changes in node embeddings will influence their corresponding distributions.
The key idea of our proposed framework is to find the most robust node embeddings, parameterized by $\theta$, that minimize the worst-case risk over the uncertainty set $\mathcal{P}_m$ in the probability simplex $\Delta_M=\{\pi \in \mathbb{R}_{+}^M: \sum_{m=1}^{M} \pi_m = 1\}$. 
This results in the definition of our distributionally robust minimax problem:
\begin{equation}
    \min _{\pi \in \Delta_M} \max _{\substack{P_m \in \mathcal{P}_m \\ 1 \leq m \leq M}} \Psi\left(\pi; P_1, \ldots, P_M\right),
    \label{eq:minmax}
\end{equation}
where $\Psi$ is the risk function of a classifier $\pi$. We define the risk function as the summation of error probabilities under each class, \ie, 
$\Psi\left(\pi ; P_1, \ldots, P_M\right):=\sum_{m=1}^M \mathbb{E}_{\xi \sim P_m}\left[1-\pi_m(\xi)\right]$. 
We note that the optimal solution $P_1^*, \dots, P_M^*$ to the inner maximization problem is known as the \emph{least favorable distributions} (LFDs) in statistics literature \cite{NEURIPS2018_a08e32d2,lehmann1986testing}.
The risk associated with these distributions is considered the worst-case risk \cite{NEURIPS2018_a08e32d2}.

\begin{figure}
  \centering
  \includegraphics[width=0.7\linewidth]{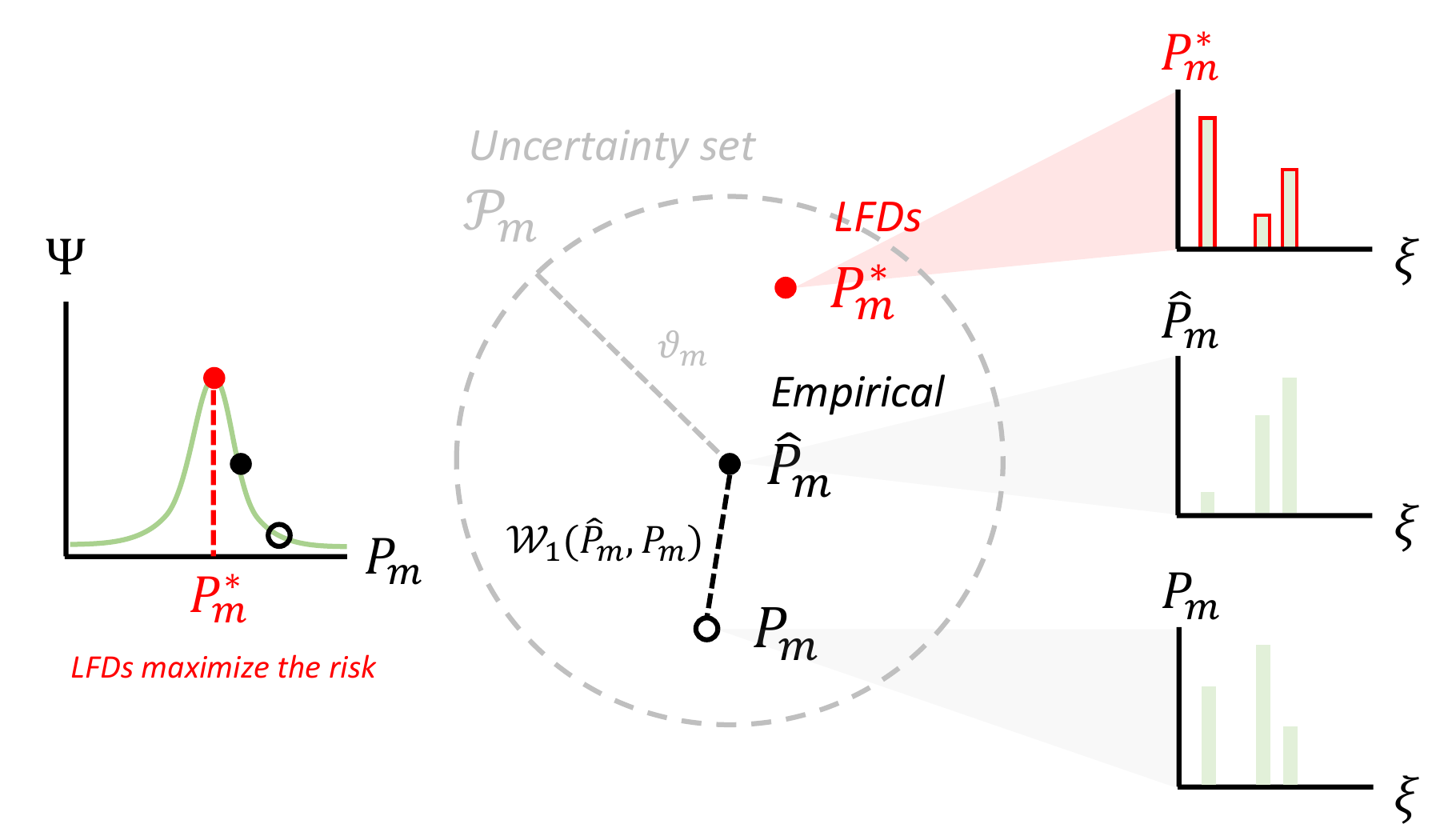}
  \caption{
  The minimax problem \eqref{eq:minmax} aims to find the least favorable distributions (LFDs) by searching the optimal $P_m$ in the uncertainty set $\mathcal{P}_m$ that maximizes the risk $\Psi$. 
  }
  \label{fig:uncertainty_set}
\end{figure}

As shown in Figure~\ref{fig:uncertainty_set}, we choose the uncertainty set $\mathcal{P}_m$ to be a Wasserstein ball of radius $\vartheta_m$ centered at the empirical distribution $\widehat{P}_m$:
\begin{equation}
    \mathcal{P}_m:=\{P_m \in \mathscr{P}(\Xi): \mathcal{W}_1(P_m, \widehat{P}_m) \leq \vartheta_m\},
    \label{eq:uncertainty-set}
\end{equation}
where $\mathscr{P}(\Xi)$ denotes the set of all probability distributions on $\Xi$. 
The Wasserstein distance of order one, $\mathcal{W}_1$, is defined as $\mathcal{W}_1\left(P, P^{\prime}\right):=\min _\gamma \mathbb{E}_{\left(\xi, \xi^{\prime}\right) \sim \gamma}\left[c\left(\xi, \xi^{\prime}\right)\right]$, where $c(u,v)$ is some cost function transferring from $u$ to $v$, $c(u,v) \ge 0$. 
The empirical distribution $\widehat{P}_m$ is represented by the Dirac point mass, denoted as:
\begin{equation}
\widehat{P}_m:=\frac{1}{\left|\left\{i: y_i=m\right\}\right|} \sum_{i=1}^n \delta_{\xi_i} \mathbbm{1}\left\{y_i=m\right\}, m=1, \ldots, M,
\label{eq:p_hat}
\end{equation}
Here, $\delta$ refers to the Dirac delta function, $|\cdot|$ represents the cardinality of a set, and $\mathbbm{1}$ denotes the indicator function.

As the original problem \eqref{eq:minmax} involves an intractable infinite-dimensional functional optimization, we follow \cite{NEURIPS2018_a08e32d2,zhu2022distributionally} and present a proposition that reformulates it into a tractable convex optimization problem. This reformulation is enabled by our careful selection of the risk function and uncertainty sets, exploiting the structure of the least favorable distributions from Wasserstein uncertainty sets \cite{NEURIPS2018_a08e32d2}.
\begin{prop}
\label{prop:1}
For the uncertainty sets defined in \eqref{eq:uncertainty-set}, the least favorable distribution of problem \eqref{eq:minmax} can be obtained by solving the following problem:
\begin{equation}
\begin{aligned}
\min _{p_1, \ldots, p_M \in \mathbb{R}_{+}^{n}} & \sum_{i=1}^n \max _{1 \leq m \leq M} p_m^i \\
\text { subject to } & \sum_{i=1}^{n} \sum_{j=1}^{n} \gamma_m^{i, j} c\left(\xi_i, \xi_j\right) \leq \vartheta_m \\
& \sum_{i=1}^{n} \gamma_m^{i, j}=\widehat{P}_m\left(\xi_j\right), \quad \sum_{j=1}^{n} \gamma_m^{i, j}=p_m^i, \\
& \forall 1 \leq i, j \leq {n}, 1 \leq m \leq M .
\label{eq:reformulation}
\end{aligned}
\end{equation}
The decision variable $\gamma_m \subset \mathbb{R}_{+}^{{n} \times {n}}$ can be viewed as a joint distribution on $n$ empirical points with marginal distributions $\widehat{P}_m$ and $P_m$, represented by a vector $p_m \in \mathbb{R}_{+}^{n}$. The inequality constraint controls the Wasserstein distance between $P_m$ and $\widehat{P}_m$.
\end{prop}
\begin{rmk}
The maximization in \eqref{eq:reformulation} 
measures the margin between the maximum likelihood of $\xi_i$ among all classes and the likelihood of the $m$-th class. Thus, the objective  can be equivalently rewritten as the minimization of the total margin.
When $M = 2$, the total margin reduces to the total variation distance.
\end{rmk}

\begin{algorithm}[t]
    \caption{Learning algorithm of \texttt{DRGL}}
    \textbf{Input}: $\mathcal{G} = (X, A)$; $y_{o} = [y_1, \dots, y_{n'}]^\top$; $\theta_0$;\\
    \textbf{Output}:  $\theta_T$
    \begin{algorithmic}[1] 
    \FOR{$t \leftarrow [0$ ... T]}
        \FOR{ \textit{each mini-set} }
        \STATE Compute the node embeddings $\{\xi\}$ given $\theta_t$ for all labeled nodes in the mini-set;
        \STATE Calculate $\widehat{P}_1, \ldots, \widehat{P}_M$ given $\{\xi\}$ and $y_o$ using \eqref{eq:p_hat};
        \STATE Obtain LFDs $P_1^*, \ldots, P_M^*$ by solving \eqref{eq:reformulation} with DO given $\{\xi\}$ and $\widehat{P}$;
        \STATE $\mathcal{L}\left(\theta_t\right) \leftarrow \mathcal{L}\left(\theta_t\right) + J\left(\theta_t ; P^*\right)$ using \eqref{eq:end-to-end loss};
        \ENDFOR
        \STATE $\theta_{t+1} \leftarrow \theta_t-\alpha \nabla \mathcal{L}\left(\theta_t\right)$ ($\alpha$ is the learning rate);
    \ENDFOR
    \end{algorithmic}
    \label{alg:drgl}
\end{algorithm}


\vspace{-.1in}
\subsection{Model Estimation}
\label{sec:model_estimation}
The proposed learning method for robust node embeddings can be carried out in an end-to-end fashion. 
To propagate the error backward through the convex optimization problem described in \eqref{eq:reformulation} to the graph encoder,
we adopt the idea of differentiable optimization (DO) \cite{cvx,amos2017optnet}.
DO enables gradient-based learning by treating the convex solver as a function that maps data to solutions. The objective is jointly formulated as:
\begin{equation}
    J\left(\theta ; P_1^*, \ldots, P_M^*\right):=\sum_{i=1}^{n'} \max _{1 \leq m \leq M} P_m^*\left(\phi_\theta(i, \mathcal{G})\right),
    \label{eq:end-to-end loss}
\end{equation}
where $P^*_m(\cdot)$ can be regarded as the output layer of our model, which takes the node embeddings $\xi_i$ as input and returns their LFDs by solving \eqref{eq:reformulation} with DO. 

We apply the mini-batch stochastic gradient descent as summarized in Algorithm~\ref{alg:drgl}.
It is necessary that each mini-batch must contain at least one training sample per class. This trivial requirement is put in place to maintain the integrity of the optimization to generate valid solutions.

\section{Experiments}
\label{sec:exp}
\subsection{Experimental Setup}
In our experiments, we use three widely-used data sets, including Cora \cite{mccallum2000automating}, Citeseer \cite{giles1998citeseer}, and Pubmed \cite{graphdata}.
These data sets consist of citation networks among $2,708$, $3,327$, and $19,717$ scientific publications, respectively \cite{yang2016revisiting}.
In these networks, each node represents a text document, and its feature vector corresponds to a bag-of-words representation.
We primarily focus on few-shot learning tasks where each data set contains $M$-class and $K$ noisy training samples per class. 

To assess the robustness of our method, we introduce random noise into these data sets in the following two ways:
    (1) We add Gaussian noise $\epsilon \sim \mathcal{N}(0, \sigma)$ to the node feature matrix $X$, where $\sigma$ is set proportionally to the standard deviation of the bag-of-words representation of all nodes in each graph;
    (2) We randomly add or remove a certain percentage $r$ of edges in the graph. 
We repeat each test three times and calculate the average accuracy.

Since our framework is designed to be compatible with various models, we choose to experiment with Graph Convolutional Networks (\texttt{GCNs}) \cite{gcn} and Graph Attention Networks (\texttt{GATs}) \cite{gat} as our graph encoders due to their popularity as competitive baselines.
We also evaluate our framework using two commonly used classifiers (output layers), including a shallow (2-layer) neural network with Softmax output and a weighted $k$-NN.
To fit the models, we use the Adam optimizer with a learning rate of $10^{-4}$ for all experiments.


\noindent\textbf{Main results.} We first evaluate the impact of adding random Gaussian noise to node features with $K=5$. As shown in Table~\ref{table:noise_result}, \texttt{GCN} and \texttt{GAT} models enhanced with \texttt{DRGL} consistently outperform their standard versions, demonstrating superior robustness against noisy node features. In the random edge perturbations setting, Table~\ref{table:combined} shows that \texttt{GCN} and \texttt{GAT}, enhanced with \texttt{DRGL}, exhibit significant improvements, particularly in the edge removal case. 
In the edge addition scenario, while \texttt{DRGL} shows a moderate improvement for \texttt{GAT}, it consistently boosts \texttt{GCN}'s performance. Additionally, in the standard few-shot setting without noise, \texttt{DRGL} improves \texttt{GCN}'s performance across all datasets, as shown in Table~\ref{table:low_data_result}, indicating its utility in both noisy and standard conditions.

\begin{table}[!t]
\centering
\caption{Model performances with Gaussian noise in node features ($K = 5$).}
\resizebox{\linewidth}{!}{%
\begin{tabular}{llllllllllll}
\toprule
Models &
\multicolumn{2}{c}{Cora (M = 7)} && \multicolumn{2}{c}{Citeseer (M = 6)} && \multicolumn{2}{c}{Pubmed (M = 3)} \\
{$\sigma$}
& {$0.112$} & {$0.224$} && {$0.091$} & {$0.182$} && {$0.018$} & {$0.036$}  \\
\midrule
\texttt{LP} & 47.70 & 47.70 && 21.73 & 21.73 && 28.90 & 28.90 \\
$\texttt{GCN+}$ Softmax & 66.17 & 52.43 && 37.83 & 30.35 &&  63.43 & 57.83 \\
$\texttt{GAT+}$ Softmax & \underline{66.48} & \underline{59.03} && \textbf{65.90} & \underline{60.43} && 63.73 & 58.70 \\
$\texttt{ProGNN}$ & 63.53 & 53.60 && 43.57 & 43.90 && OOM & OOM \\
$\texttt{RGCN}$ &  61.43 & 52.87 && 45.93 & 36.13 && \underline{64.10} & 59.03 \\
$\texttt{GCN-SVD}$ &  55.27 & 55.27 && 33.57 & 31.53  && 49.80 & 49.80 \\
\midrule
\textbf{$\texttt{GCN}_{\texttt{DRGL}} +$ Softmax} & 66.20 & 54.40 && 39.80 & 34.90 && 63.20 & \underline{59.70} \\
\textbf{$\texttt{GAT}_{\texttt{DRGL}} +$ Softmax} & \textbf{67.83} & \textbf{59.40} && \underline{65.60} & \textbf{61.27} && \textbf{64.57} & \textbf{59.77} \\

\bottomrule
\end{tabular}
}
\label{table:noise_result}
\begin{tablenotes}
      \small
      \item * $\sigma$ takes value of one or two standard deviation of the nodal features.
\end{tablenotes}
\vspace{-.1in}
\end{table}

\begin{table}[!t]
\begin{center}
\caption{Combined Model Performances with random edge removal and addition ($K = 5$)}
\vspace{-.09in}
\resizebox{\linewidth}{!}{%
\begin{tabular}{lcccccc}
\toprule
{Models} & \multicolumn{2}{c}{Cora (M = 7)} & \multicolumn{2}{c}{Citeseer (M = 6)} & \multicolumn{2}{c}{Pubmed (M = 3)} \\
{Modification ($r$)} & {$\pm20\%$} & {$\pm50\%$} & {$\pm20\%$} & {$\pm50\%$} & {$\pm20\%$} & {$\pm50\%$} \\
\midrule
\texttt{LP} & 41.67/47.70 & 32.10/47.70 & 17.83/21.73 & 11.60/21.73 & 26.53/28.90 & 23.53/28.90 \\
$\texttt{GCN+}$ Softmax & \underline{67.30}/61.73 & 50.83/55.03 & 39.25/\textbf{47.67} & 35.70/39.83 & \underline{66.07}/59.27 & \underline{62.07}/\textbf{60.07} \\
$\texttt{GAT+}$ Softmax & 67.28/\textbf{68.93} & \underline{57.08}/\textbf{63.00} & \underline{64.07}/47.20 & \underline{60.13}/\textbf{44.27} & 64.80/\textbf{63.70} & 61.96/59.53 \\

$\texttt{ProGNN}$ & 63.70/66.63 & 56.60/56.60 & 46.73/43.00 & 46.73/38.00 & OOM/OOM & OOM/OOM \\
$\texttt{RGCN}$ & 62.76/57.23 & 56.33/50.97 & 47.00/42.63 & 43.77/39.20 & 63.93/61.76 & 58.20/56.00 \\
$\texttt{GCN-SVD}$ & 32.53/47.80 & 27.80/44.37 & 32.53/32.06 & 27.80/31.30 & 47.40/51.90 & 44.37/50.07 \\
\midrule
\textbf{$\texttt{GCN}_{\texttt{DRGL}}+$ Softmax} & 66.70/62.63 & 52.15/55.90 & 40.15/\textbf{47.67} & 41.80/40.26 & \textbf{66.63}/60.56 & \textbf{64.30}/\underline{59.80} \\
\textbf{$\texttt{GAT}_{\texttt{DRGL}}+$ Softmax} & \textbf{69.02}/\underline{67.60} & \textbf{58.88}/\underline{59.13} & \textbf{65.93}/\underline{47.30} & \textbf{61.20}/\underline{43.97} & \underline{65.93}/\underline{62.30} & 60.53/59.03 \\
\bottomrule
\end{tabular}
\label{table:combined}
}
\end{center}
\begin{tablenotes}
      \small
      \item * $r$ represents the percentage of edges either removed (-) or added (+). Performance scores for removal/addition are separated by '/' for clarity.
\end{tablenotes}
\vspace{-.2in}
\end{table}

\subsection{Learned Embeddings and Uncertainty} To gain a more intuitive understanding of the learned embedding space produced by \texttt{DRGL}, we conducted an additional ablation study. We use a two-dimensional embedding space
and visualize it as a scatter plot. In these figures, large dots represent training points, and small dots represent testing points, colored by true categories. The depth of color indicates the likelihood of classification.

\begin{figure}[!t]
  \centering
  \subfigure[Raw (\texttt{DRGL})]
  {\includegraphics[width=0.14\textwidth]{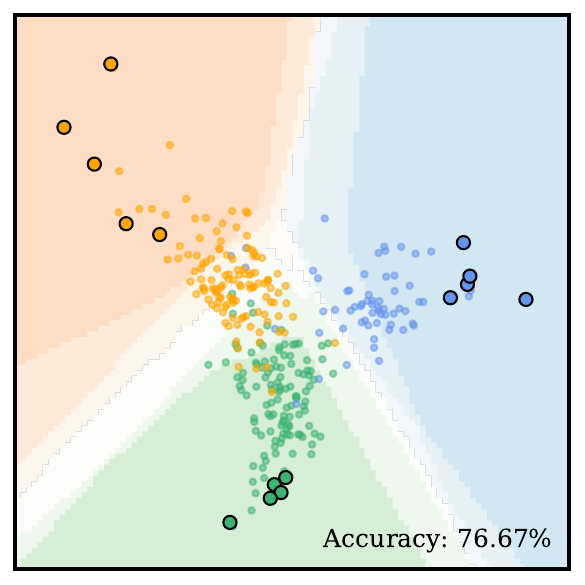}}
  \subfigure[$2\sigma$ noise (\texttt{DRGL})]{\includegraphics[width=0.14\textwidth]{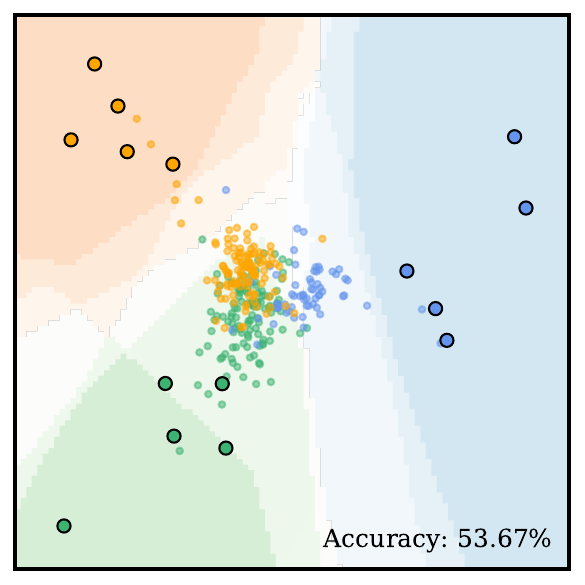}}
  \subfigure[20\% edge (\texttt{DRGL})]{\includegraphics[width=0.14\textwidth]{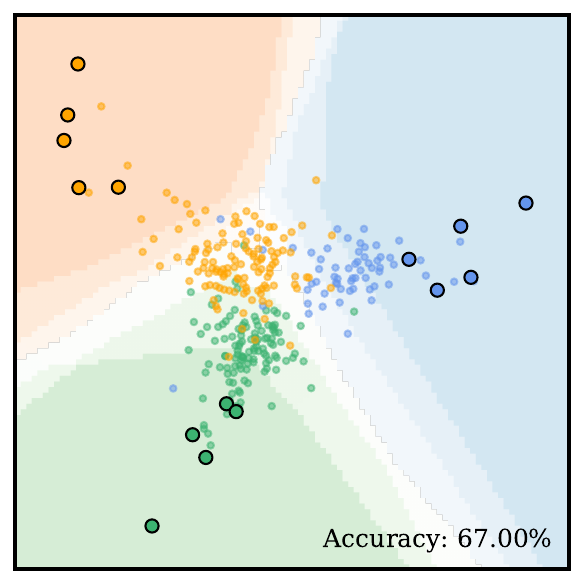}}
  \\
\subfigure[Raw (\texttt{GCN})]{\includegraphics[width=0.14\textwidth]{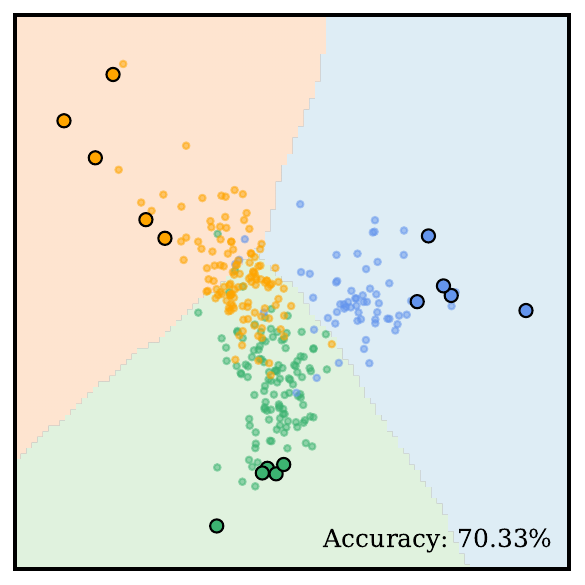}}
  \subfigure[$2\sigma$ noise (\texttt{GCN})]{\includegraphics[width=0.14\textwidth]{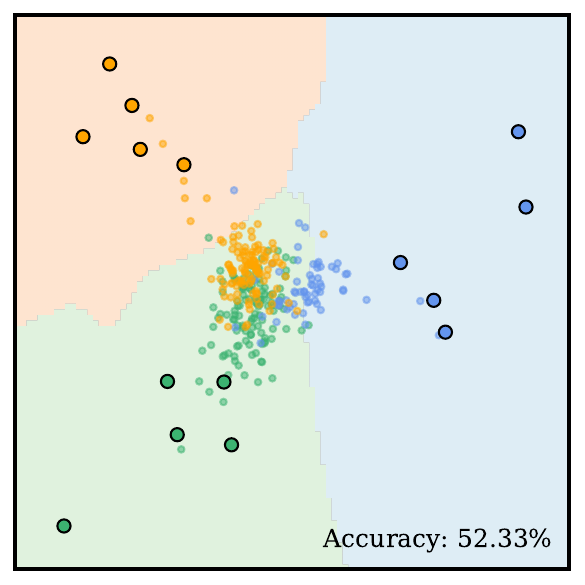}}
  \subfigure[20\% edge (\texttt{GCN})]{\includegraphics[width=0.14\textwidth]{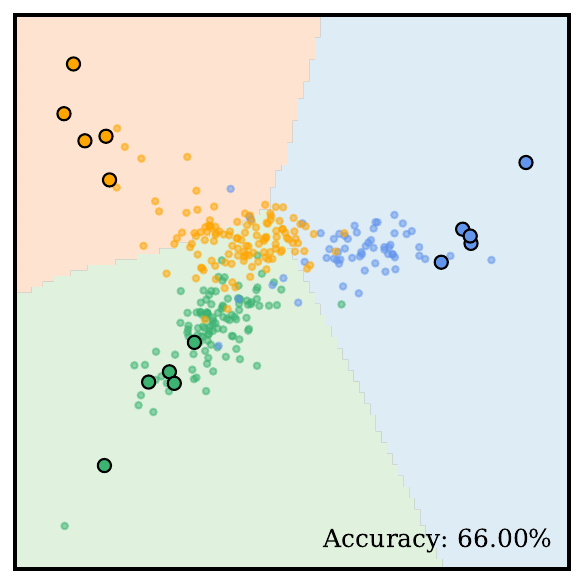}}
  \caption{The impact of noise on the learned feature spaces. (a) and (b) show the embeddings from graphs without noise; (b) and (e) show the embeddings when the graphs have nodal features with $2\sigma$ noise; and (c) and (f) present the representations from graphs where $20\%$ of the edges have been removed.
  }
  \label{fig:hidden_rep}
  \vspace{-.1in}
\end{figure}

\begin{figure}[!t]
\centering
\vspace{-.1in}
\subfigure[\texttt{DRGL} + \texttt{GCN}]{\includegraphics[width=0.4\linewidth]{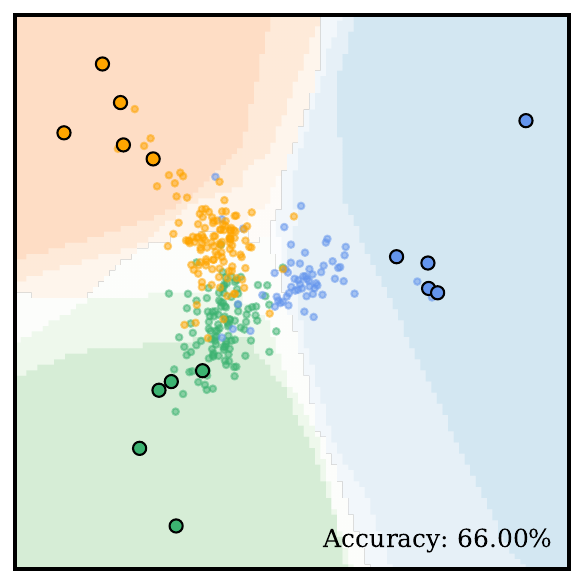}}
\subfigure[UQ of \texttt{DRGL} + \texttt{GCN}
]{\includegraphics[width=0.4\linewidth]{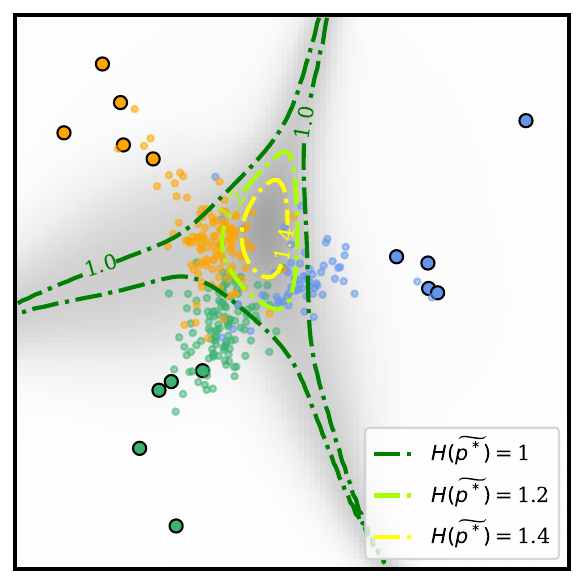}}
\caption{The learned embeddings and the uncertainty of GCN trained with \texttt{DRGL} Darker
shades indicate a higher level of uncertainty between the different categories under that LFDs solved by \eqref{eq:reformulation}. 
}
\label{fig:lfd_comparison}
\vspace{-.1in}
\end{figure}

Figure \ref{fig:hidden_rep} compares the embeddings from \texttt{DRGL} and vanilla \texttt{GCN}, showing slightly increased between-class separation and reduced intra-class distance. This subtle change significantly improves classification accuracy.
Figure \ref{fig:lfd_comparison} visualizes the LFDs using kernel density estimation, with darker regions indicating higher uncertainty. The predictive uncertainty can be expressed using entropy 
$
H(\tilde{p}^*)=-\sum_{m=1}^{M} \tilde{p}_m^* \log \tilde{p}_m^*
$ \cite{6769090, namdari2019review, schwill2018entropy}.
The result highlights the improved capability of our approach in capturing and quantifying uncertainty compared with the vanilla methods.

\begin{table}[!t]
\centering
\caption{Model performances without graph noise ($K=5$ and $K = 10$).}
\resizebox{\linewidth}{!}{%
\begin{tabular}{llllllllllll}
\toprule
\multirow{2}{*}{Models} &
\multicolumn{2}{c}{Cora (M = 7)} && \multicolumn{2}{c}{Citeseer (M = 6)} && \multicolumn{2}{c}{Pubmed (M = 3)} \\
& {K = 5} & {K = 10} && {K = 5} & {K = 10} && {K = 5} & {K = 10}  \\
\midrule
\texttt{LP} & 47.70 & 52.80 && 21.73 & 28.33 && 28.90 & 38.63 \\
$\texttt{GCN+}$ $k$-NN & 36.20 & 66.50 && 28.50 & 52.47 && 45.57 & 68.27 \\
$\texttt{GCN+}$ Softmax & 63.47 & 70.67 && 45.37 & \underline{57.83} && 67.10 & 71.27 \\
$\texttt{ProGNN}$ & \textbf{71.20} & \textbf{75.80} && 45.10 & 54.53 && \underline{67.70} & OOM \\
$\texttt{RGCN}$ & \underline{66.13} & 71.23 && \textbf{50.00} & 53.00  && 66.43 & 70.53 \\
$\texttt{GCN-SVD}$ & 52.67 & 60.43 && 33.57 & 34.34 && 52.17 & 57.43 \\
\midrule
$\texttt{GCN}_{\texttt{DRGL}}+$ $k$-NN & 44.60 & 68.23 && 34.07 & 53.30 && 51.10 & 68.67 \\
$\texttt{GCN}_{\texttt{DRGL}}+$ Softmax & \underline{66.13} & \underline{72.60} && \underline{49.83} & \textbf{59.33} && \textbf{67.83} & \textbf{72.20} \\
\bottomrule
\end{tabular}
}
\label{table:low_data_result}
\vspace{-.1in}
\end{table}

\section{Conclusion}
\label{sec:conlusion}
To address the challenges posed by noisy graphs, we proposed a novel GNN-agnostic framework that enhances the robustness of node embeddings and predictive performance. Specifically, our framework improves model robustness by accounting for uncertainties introduced by data noise in the graph. This approach demonstrates substantial improvements over state-of-the-art baselines across various benchmark datasets.

\bibliographystyle{unsrt}  
\bibliography{ref}

\end{document}